\def\year{2020}\relax
\documentclass[letterpaper]{article} 
\usepackage{aaai20}  
\usepackage{times}  
\usepackage{helvet} 
\usepackage{courier}  
\usepackage[hyphens]{url}  
\usepackage{graphicx} 
\usepackage{graphicx}  
\usepackage{comment}
\usepackage{amsmath}
\usepackage{multirow}
\usepackage{graphicx}
\usepackage{subfigure}
\title{Endowing Deep 3D Models with Rotation Invariance \\
	Based on Principal Component Analysis}
\author{Zelin Xiao$^1$$^,$$^2$$^,$,
	Hongxin Lin$^1$$^,$$^2$$^,$,	
	Renjie Li$^1$$^,$$^2$,
	Hongyang Chao$^1$,
	Shengyong Ding$^1$$^,$$^*$ \\
	$^{1}$School of Data and Computer Science, Sun Yat-sen University, Guangzhou, China \\
	$^{2}$Pixtalks Tech, Guangzhou, China \\
	\{xiaozl,linhx9, lirj25\}@mail2.sysu.edu.cn, isschhy@mail.sysu.edu.cn, marcding@163.com}

\begin{document}

\maketitle

\begin{abstract}
In this paper, we propose a simple yet effective method to endow deep 3D models with rotation invariance by expressing the coordinates in an intrinsic frame determined by the object shape itself. Key to our approach is to find such an intrinsic frame which should be unique to the identical object shape and consistent across different instances of the same category, e.g. the frame axes of desks should be all roughly along the edges. Interestingly, the principal component analysis exactly provides an effective way to define such a frame, i.e. setting the principal components as the frame axes. As the principal components have direction ambiguity caused by the sign-ambiguity of eigenvector computation, there exist several intrinsic frames for each object. In order to achieve absolute rotation invariance for a deep model, we adopt the coordinates expressed in all intrinsic frames as inputs to obtain multiple output features, which will be further aggregated as a final feature via a self-attention module. Our method is theoretically rotation-invariant and can be flexibly embedded into the current network architectures. Comprehensive experiments demonstrate that our approach can achieve near state-of-the-art performance on rotation-augmented dataset for ModelNet40 classification and outperform other models on SHREC’17 perturbed retrieval task.
\end{abstract}

\section{Introduction}

With the development of 3D sensors such as structured light, time-of-flight and LIDAR, 3D data can be easily acquired and directly processed in many applications, such as autonomous driving, 3D face recognition and etc. In general, 3D data is encoded in the form of point cloud, which directly records the coordinates of the sampled points on the object surface. A key challenge for point cloud processing is that the input data is highly rotation-variant, which means a 3D object possesses rotated clones in infinite attitudes. This remains an intractable problem even for recently proposed deep 3D models such as PointNet \cite {qi2017pointnet}, PointNet++ \cite{qi2017pointnet++} and DGCNN \cite{wang2018dynamic}.

To alleviate the rotation variance problem, typical approaches  either use a spatial transformer module as in the original PointNet \cite{qi2017pointnet}  or apply extensive data augmentation during the training phase. However, it requires higher model capacity and brings extra computation burdens. Other methods such as Spherical CNN \cite{esteves2018learning} and SFCNN \cite{rao2019spherical} focus on converting the point cloud into some special structures to extract a  rotation-invariant feature, which might suffer from loss of information.
\begin{figure}[tp] \centering
	\includegraphics[width = 0.46 \textwidth,height = 0.24 \textheight]{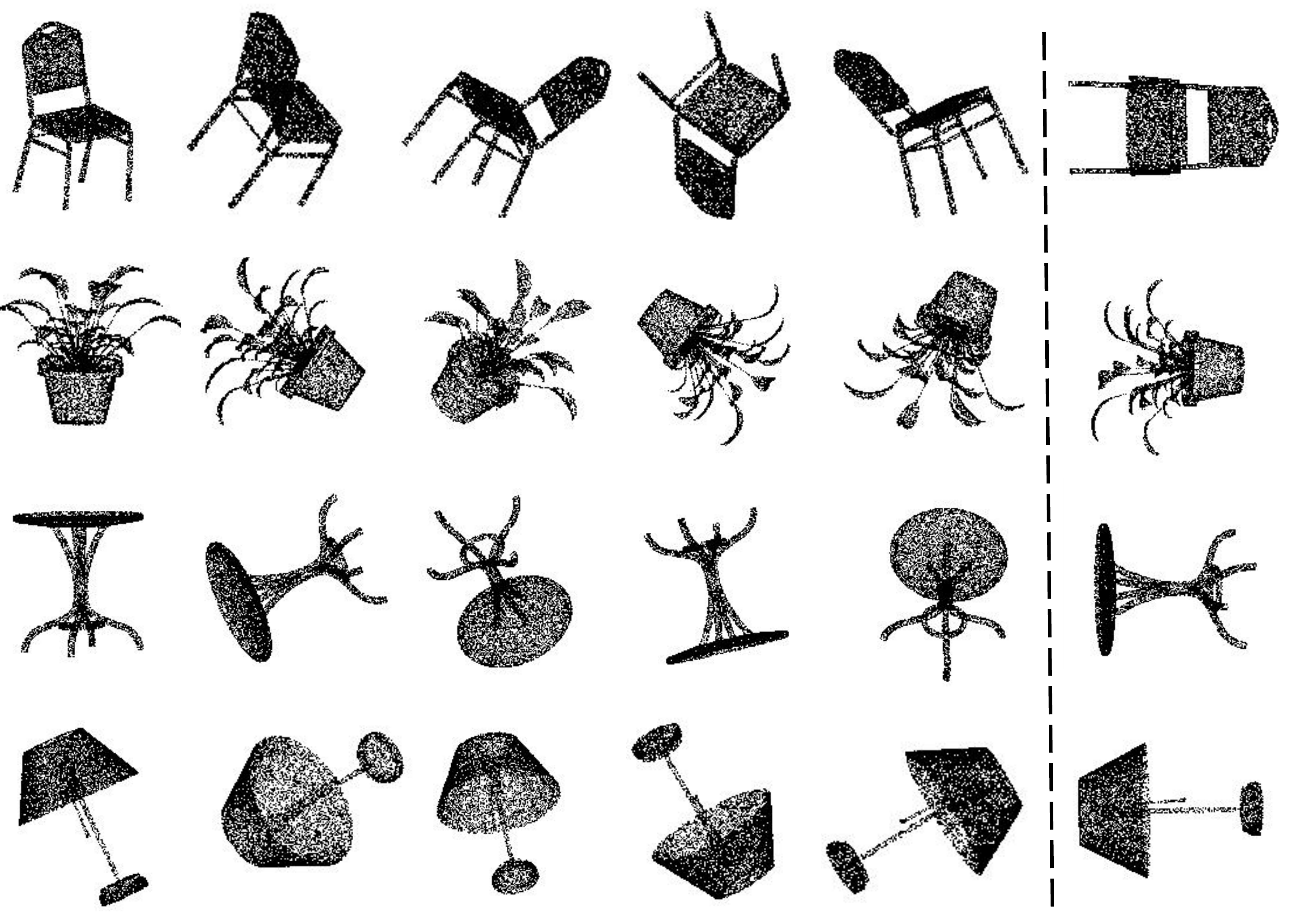}
	\caption{Illustration of our PCA-RI Representation. For each row, the left side of the dotted line are  point clouds of the identical object in different poses. All these point clouds are transformed to the consistent PCA-RI representation as shown in the right side by our method. }
\end{figure}

In this paper, we introduce a novel PCA-RI (PCA Rotation-Invariant) representation to endow deep 3D models with rotation invariance by expressing the point cloud in the intrinsic frame. Such a frame should be stable regardless of arbitrary rotations. In other words, the expressed coordinates don't change no matter how the object rotates. Besides, the intrinsic frame should be able to tolerate small distortions, thus providing a consistent representation for similar objects. Recall that PCA (principal component analysis) is designed to detect the main directions along which the variance is large for high-dimensional input data. These directions encode the intrinsic structure of the input data and maintain absolutely rotation-equivariant, which exactly offers an effective way to define the desired frames.

More specifically, we apply PCA techniques to obtain  three principal components of a point cloud, which is used as the corresponding $x, y, z$ axis of the intrinsic frame. After that, we project the point cloud onto the new frame and use the transformed coordinates as our PCA-RI representation for the point cloud, which is shown in Figure 1. A complete rotation-invariant ability will come immediately with this  method as can be proven in later sections. Compared with the previous works, our PCA-RI representation has the advantages of simplicity and generality. It can be flexibly embedded into the current deep neural networks to fundamentally improve their robustness against rotation transformation.

One concern with our proposed approach is that we are not certain of the direction of each principal component. Thus for a point cloud, there exist two directions for each frame axis, which we call frame ambiguity. To address the problem, we propose a multi-frame approach to enumerate all the possible frames derived from the principal component analysis algorithm. After that, we feed all the PCA-RI representations of these frames to the deep model and aggregate all the output features via a self-attention module. In the end, we apply an average-pooling operation after the self-attention module to extract a final feature vector for downstream tasks. To empirically validate the effectiveness of our method, we conduct a comprehensive experimental study on ModelNet40 \cite{wu20153d} classification and SHREC’17 \cite{savva2017large} perturbed retrieval tasks. The experimental results demonstrate that our approach can achieve near state-of-the-art performance on rotation-augmented dataset for ModelNet40 \cite{wu20153d} classification and outperform other models on SHREC’17 \cite{savva2017large} perturbed retrieval task.

In summary, the \textbf{key contributions} of this paper are as follows:
\begin{itemize}
	\item We propose a theoretically rotation-invariant and absolutely information-lossless point cloud representation.

	\item We further introduce a multi-frame approach based on a  self-attention module, which can effectively address the problem of frame ambiguity. 
	
	\item Extensive experiments further demonstrate the correctness and effectiveness of our method. 
\end{itemize}


\section{Related Work}

\subsection{Deep Learning for 3D Objects}
Motivated by the breakthrough results of convolutional neural networks in 2D images, increasing attention has been drawn to developing such methods for geometric data. One intuitive idea is to convert irregular point clouds into regular 3D grids by voxelization \cite{maturana2015voxnet,qi2016volumetric},  since its format is similar to pixel and easy to transfer to existing frameworks. 
However, it is inevitable to suffer from loss of resolution and high computational demand. 
To avoid the shortcoming of naive voxelization, kd-tree \cite{klokov2017escape} and octree \cite{riegler2017octnet} based methods hierarchically partition space to exploit  input sparsity. But these methods focus more on subdivision of a volume rather than local geometric structure. 

An important architectural model that directly processes point cloud is PointNet \cite{qi2017pointnet}, which adopts spatial transform networks and a symmetry function to maintain the invariance of permutation.  
After that, many point-based learning approaches focus on how to efficiently capture local features based on PointNet \cite{qi2017pointnet}. For instance, PointNet++ \cite{qi2017pointnet++} applies PointNet \cite{qi2017pointnet} structure in local point sets
with different resolutions and accumulates local features in a hierarchical architecture. 
In DGCNN \cite{wang2018dynamic}, EdgeConv is proposed as a basic block to build networks, in which the edge features between points and their neighbors are exploited.

\subsection{Self-Attention}
Recently, attention mechanisms \cite{bahdanau2014neural,show2015tell,gregor2015draw,yang2016stacked,chen2017pixelsnail} have become an integral part of models that must capture global dependencies. In particular, self-attention \cite{cheng2016long,parikh2016decomposable,vaswani2017attention}, also called intra-attention, exhibits a better balance between the ability to model long-range dependencies and the computational efficiency. The self-attention module calculates response at a position as a weighted sum of the features at all positions, where the weights called attention vectors are calculated with only a small computational cost. Vaswani et al. \cite{vaswani2017attention} further
demonstrate that machine translation models could achieve state-of-the-art results by solely using a self-attention model.
\subsection{Rotation-Invariant Network for 3D Objects}
The rotation robustness is essential in real-world applications of point cloud processing systems. Previous works have attempted to equip the existing neural networks with the property of rotation invariance. A straightforward method is to train a deep model with great amounts of rotation-augmented data. Although data augmentation is effective to some extent, it is computationally expensive during the training phase. Furthermore, the previous study \cite{esteves2018learning} has shown that aggressive data augmentation like arbitrary 3D rotations on input data will still harm the recognition performance. PointNet \cite{qi2017pointnet} applies spatial transformer network (STN) to canonicalize the input data but further experiments demonstrate that model with STN still suffers from great performance drop on arbitrary rotation-augmented 3D dataset. 

In closely related works, Esteves et al. \cite{esteves2018learning} propose a special convolutional operation with local rotation invariance, which can generalize well to unseen rotations. Besides, Rao et al. \cite{rao2019spherical} design a trainable neural network to project the original points onto the fractal structure adaptively, which makes their model resistant to arbitrary rotations. While the theoretical foundations of these approaches are well-studied, they have primarily been applied to spherical shapes  or projected  onto a structure, which might suffer from loss of information. ClusterNet \cite{chen2019clusternet} introduces a point cloud representation by using rigorously rotation-invariant operator such as the inner product between points. Although ClusterNet \cite{chen2019clusternet} claims their representation is conditional information-lossless, their experiments conducted on ModelNet40 \cite{wu20153d}  reveal that this representation still degrades the  performance on 3D tasks.

\section{Approach} \label{method}

In this section, we first introduce our PCA-RI representation for point cloud based on principal component analysis. Then we explain how to address the problem of frame ambiguity in a deep neural network by multi-frame fusion based on a self-attention module. In the end, we present how our method can be embedded into deep 3D models.
\subsection{PCA-RI Representation}
The main idea of our method is to find an intrinsic frame determined by the object shape. The intrinsic frame should provide the same representations for all rotated clones of the identical object. In addition, it should be capable of tolerating small distortion of the object shape. That is, similar objects will provide similar frames and representations. To this end, we propose a PCA-RI (PCA Rotation Invariant) representation based on the classical PCA (principal component analysis).

Let $P = \{p_i = (x_i,y_i,z_i)^ T | i = 1, ..., n  \}$ represents a point cloud, which directly encodes the coordinates of the sampled points on the object surface.  Note that the coordinate value of each point depends on the selection of the coordinate system, namely the frame. The intrinsic frame is such a frame that can be automatically detected from the object structure.

Use $\bar{p}$ to denote the mean of a point cloud $P$ and $\Sigma$ to denote its corresponding covariance matrix, which is a semi-definite symmetric matrix. Then $\Sigma$ can be calculated as follows:
\begin{eqnarray}
\Sigma = \frac{1}{n}\sum_{i=1}^{n}(p_i-\bar{p})(p_i-\bar{p})^T
\end{eqnarray}

Then we use eigendecomposition to find the eigenvector $u$ of the covariance matrix $\Sigma$, which satisfies the following equation:
\begin{eqnarray}
\Sigma u = \lambda u
\end{eqnarray}

Obviously, there are three eigenvalues, denoted as $\lambda_1,\lambda_2,\lambda_3$ with three unit normalized corresponding eigenvectors $u_1, u_2, u_3$. After that, we use $ U=[u_1,u_2,u_3]$ to define the intrinsic frame and express the point cloud $P$ in the new frame with the order $\lambda_1>\lambda_2>\lambda_3$:  
\begin{eqnarray}
p'_i=U^T(p_i-\bar{p})
\end{eqnarray}
in which $p'_i$ represents the redefined coordinate value in the intrinsic frame.
Now we will prove that the intrinsic coordinate of $p'_i$  will not change with rigid rotations. 

\begin{figure}[t] \centering
	\includegraphics[width = 0.46 \textwidth,height = 0.32 \textheight]{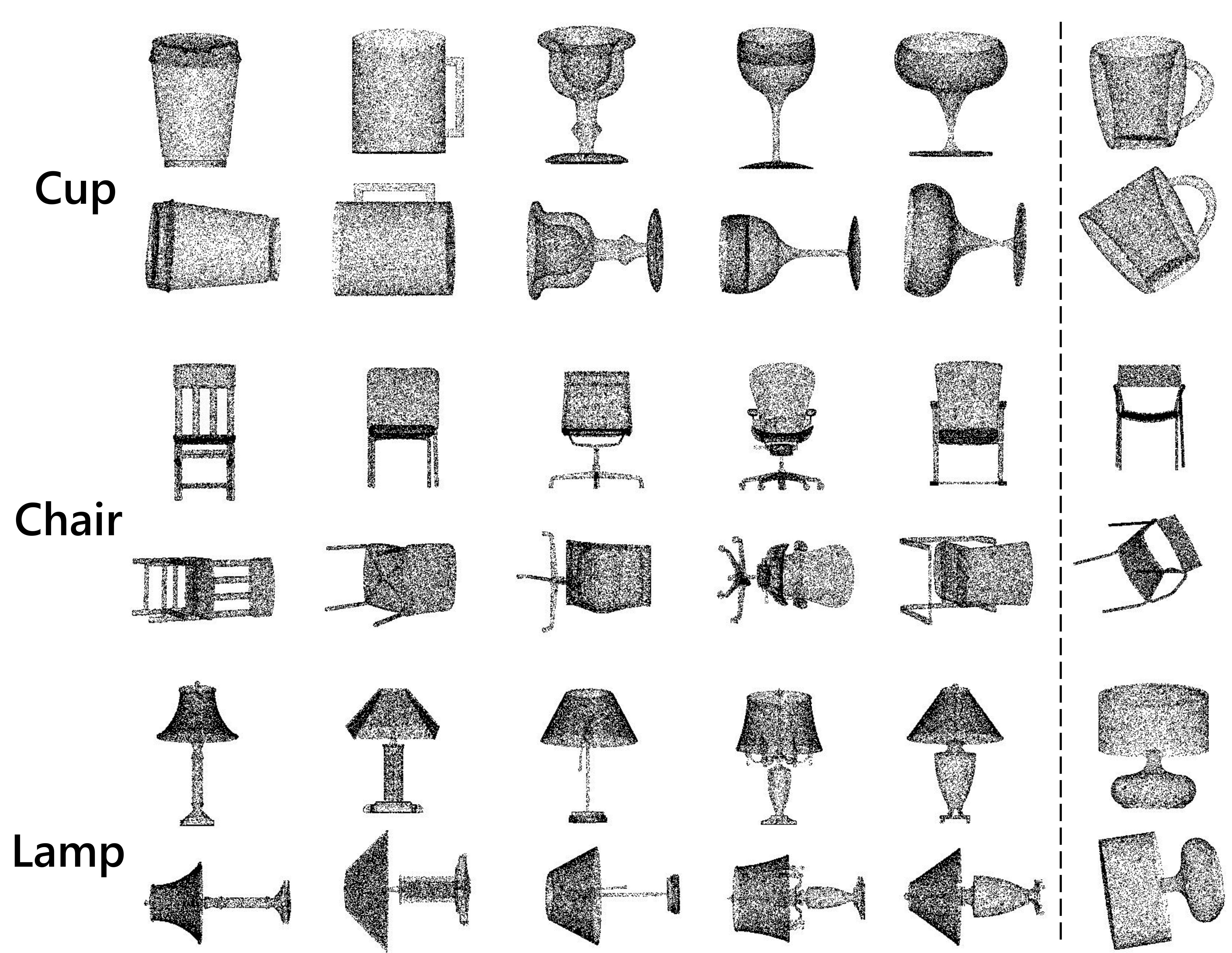}
	\caption{Comparisons between intrinsic frame alignment and manual alignment. For each category, the first row denotes the manually aligned objects  while the second row represents the objects aligned with our intrinsic frames. The last column provides some samples that cannot be perfectly aligned with our intrinsic frames.}
\end{figure}

Suppose the point cloud $P$ is rotated in the original frame, giving another rotated point cloud representation  $\hat P$ with $\hat p_i = R p_i$ where $R$ represents a rigid rotation matrix. It's not hard to see that the corresponding covariance matrix $\hat \Sigma$ of $\hat P$ satisfies 
\begin{eqnarray} 
\hat \Sigma = R\Sigma R^T
\end{eqnarray}
Obviously, we have:
\begin{eqnarray} 
\begin{split}
\hat \Sigma Ru &= R\Sigma R^T Ru \\
&= \lambda Ru
\end{split}
\end{eqnarray}
which means $\lambda_1,\lambda_2,\lambda_3$ and $\hat U = [Ru_1,Ru_2,Ru_3]$ are the eigenvalues and eigenvectors of $\hat \Sigma$ respectively. Denote  the redefined coordinate of $\hat p_i$ in the intrinsic frame as $\hat p'_i$. Thus we have:
\begin{eqnarray} 
\begin{split}
\hat p'_i &= \hat U^T (\hat p_i - \hat {\bar{p}} )\\
&= [Ru_1,Ru_2,Ru_3]^TR(p_i-\bar{p}) \\
&= U^T(p_i-\bar{p}) \\
&= p'_i  
\end{split}
\end{eqnarray}

As shown in Equation (6), the redefined coordinate value of each point in the intrinsic frame remains invariant no matter how the point cloud is rotated. From geometric perspective, our PCA-RI representation merely adjusts the arbitrary rotated clones of the identical point cloud to a consistent pose. It reveals that our PCA-RI representation is absolutely information-lossless. Apart from this, our approach is very general and can be applied to the current neural network architectures.

Note that the above rotation invariance assumes that there are three distinct eigenvalues so that we can define $x, y, z$ axis according to the order of $\lambda_1,\lambda_2,\lambda_3$, which we call axis significance. If the axis significance is weak, i.e. the eigenvalues are close, we will not be able to detect a stable intrinsic frame. For instance, if the shape is composed of three intersected orthogonal lines of the same length, then the covariance matrix is an identity matrix $I$ and $\lambda_1=\lambda_2=\lambda_3$. This will result in infinite intrinsic frames and meanwhile the property of rotation invariance  will not hold anymore. Fortunately, the axis significance will be preserved in general cases as shown in the later experiments.

Another noteworthy point is whether the intrinsic frames are consistent across intra-class objects. For example, the axes of intrinsic frames for desks are all roughly along the edges. We argue that for the same category, the principal components are close, thus ensuring the frames consistent. As shown in Figure 2, we list some examples for cup, chair and lamp categories. For each category, the first row denotes the manually aligned objects while the second row represents the objects aligned with our intrinsic frames. It's not hard to see that the intrinsic frames are consistent across the intra-class samples in most cases. Note that we cannot  still achieve absolute alignment like manual alignment for the reason that PCA pays more attention to  the data distribution of point cloud. Despite this,  theoretical analysis   and extensive experiments still demonstrate that the canonicalization of our PCA-RI representation can essentially reduce the learning difficulty of the neural network by  replacing infinite rotating attitudes with some fixed poses and meanwhile retaining the original point cloud information intact. 

\subsection{Frame Ambiguity Elimination}
One concern with our proposed method is that when we try to define the new coordinates using eigendecomposition, we are not sure about the direction of $u$ as
the following equation (7)  also holds.
\begin{eqnarray} 
\Sigma (-u) = \lambda (-u)
\end{eqnarray}

Specifically, the process of eigenvector computation provides no means for assessing the sign of each eigenvector so that the individual eigenvector has an arbitrary sign. It means that for the identical point cloud, there exist two directions for each frame axis, which we denote as frame ambiguity. Figure 3 illustrates the phenomenon of frame ambiguity.

To address the issue, we adopt a multi-frame approach to fuse the results. Denoting the deep model which we are going to endow with rotation invariance by a function $Y=F(P)$. The $F(P)$ denotes a feature vector generated by the deep model when given the input point cloud $P$. Here we suppose the centroid of the input point cloud is on the origin. Our fusion scheme can be abstracted as follows:
\begin{equation}
Y_{fused} = h\{F(U_1^{T}P),F(U_2^{T}P),...,F(U_8^{T}P)\}
\end{equation}
in which we introduce a fusion function $h$ to obtain a final feature descriptor $Y_{fused}$ from multiple PCA-RI representations $U_i^{T}P$ with $U_i$ denoting the $i^{th}$ frame. 

In order to achieve absolute rotation invariance, we require the fusion function $h$ to be independent on the order of the frames with $F_i$ = $F(U_i^{T}P)$ and $\pi$ denoting the permutation respectively as follows:
\begin{equation}
h(F_1,F_2,...,F_8)=h(F_{\pi(1)},F_{\pi(2)},...,F_{\pi(8)}) 
\end{equation}

A naive approach is to directly apply an average or max pooling operation on $\{F_i\}$. However, we find that the direct pooling operation disregards a lot of relationship among the features, which limits the discriminability of the final feature.

\begin{figure}[t] \centering
	\includegraphics[width = 0.48 \textwidth,height = 0.15 \textheight]{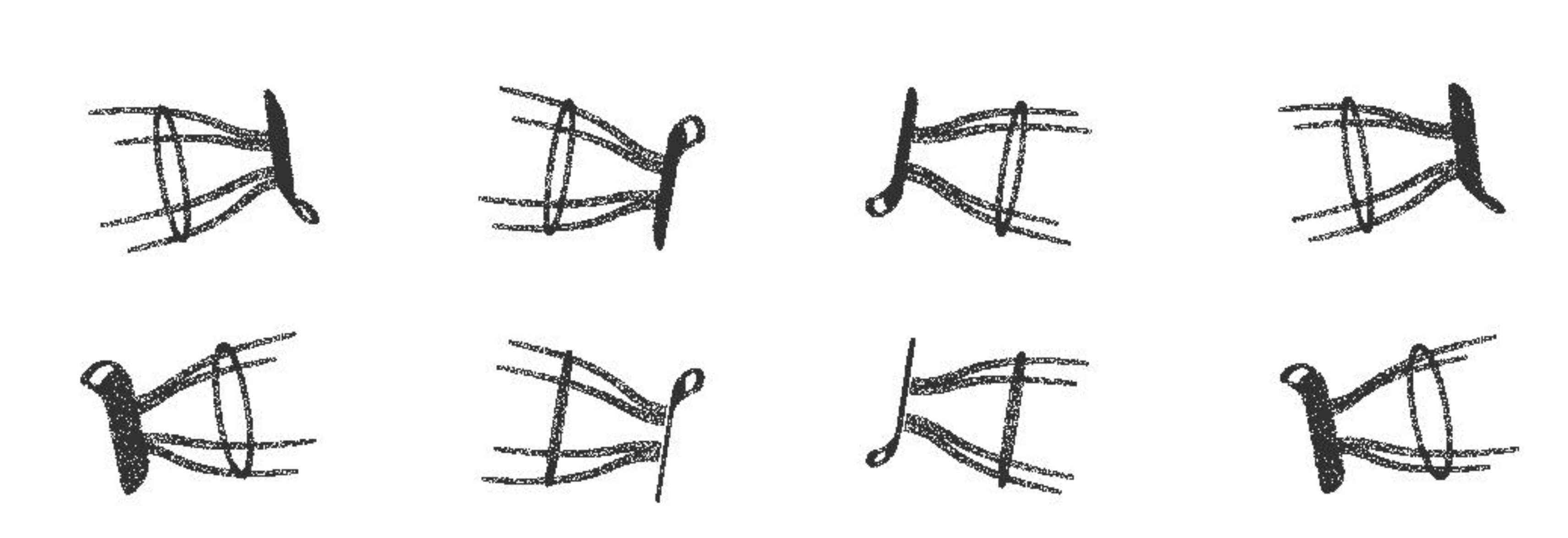}
	\caption{Illustration of Frame Ambiguity. There are eight possible intrinsic frames in total for each scan. We show the projected points onto the $XY$ plane for better visualization.}
\end{figure}
\begin{figure*}[tb] \centering
	\includegraphics[width = \textwidth,height = 0.32 \textheight]{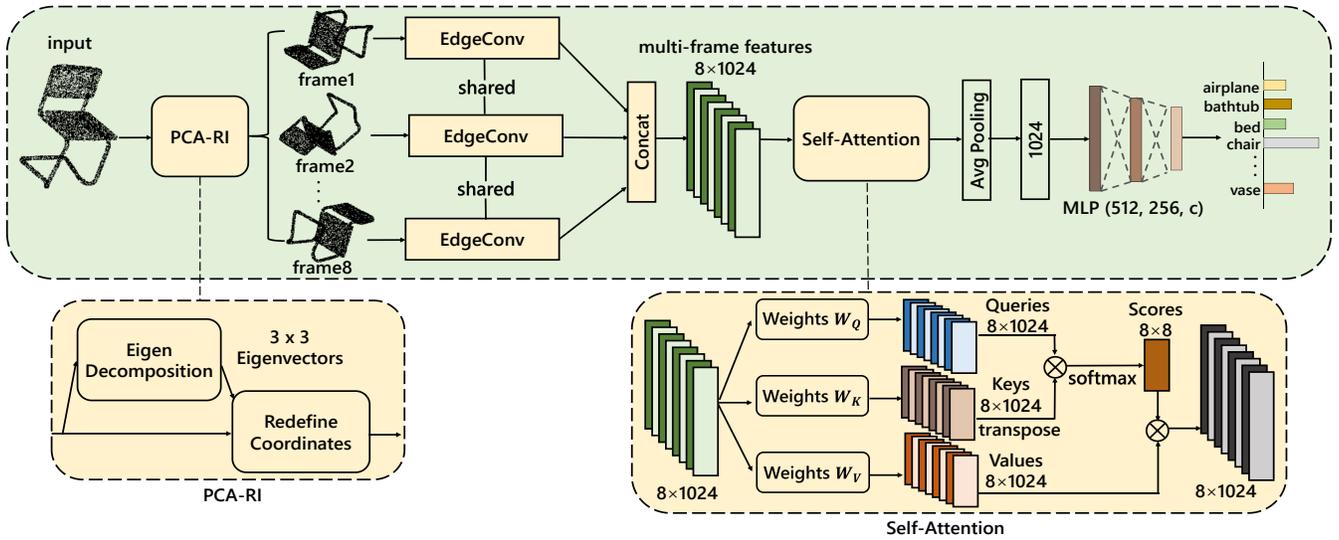}
	\caption{Overview of our architecture. Our proposed  method can be easily extended to various 3D tasks  including classification and retrieval. In our framework, we first convert the point cloud into the PCA-RI representation. Then, a shared five-layer EdgeConv module with layer output sizes 64, 64, 64, 128, 1024 respectively is used to capture local and global patterns of point cloud and thus we obtain eight 1024-dim features in total for all intrinsic frames. After that, a self-attention module is applied to  indicate how much attention is paid on each frame. Finally, we use an average pooling layer after the self-attention module to obtain a final feature for the downstream tasks.}
\end{figure*}

To alleviate the problem, we apply a self-attention module derived from \cite{vaswani2017attention} before pooling layers to pay more attention to the relationship between the multi-frame features.  Following the notations of \cite{vaswani2017attention}, the transformed feature $\hat F_i$ derived from the self-attention module can be expressed as follows with shared parameter matrices $W_Q$, $W_K$ and $W_V$:
\begin{eqnarray}
\begin{split}
\hat F_i &= \Sigma_{j}\frac{\exp{<F_iW_Q,F_jW_K>}}{\Sigma_{l}\exp{<F_iW_Q,F_lW_K>}}F_jW_V \\
\end{split}
\end{eqnarray}

From Equation (10) we can observe that the attention module aims to allocate weight to multi-frame feature $F_i$ and accumulate the weighted features. 

Use $G$ to denote the self-attention transformation as follows:
\begin{equation}
(\hat F_{1},\hat F_{2},...,\hat F_{8}) = G(F_{1},F_{2},...,F_{8})
\end{equation}

We care about whether the $\{\hat F_i\}$ is invariant to the input order of $F_i$ as Equation (12) shows:
\begin{equation}
(\hat F_{\pi(1)},\hat F_{\pi(2)},...,\hat F_{\pi(8)}) = G(F_{\pi(1)},F_{\pi(2)},...,F_{\pi(8)})
\end{equation}

Fortunately, this equation holds as the sum operation in Equation (10) doesn't care about the order of items. In summary, our transformed features derived from the self-attention module are independent on the order of input frames.

With these transformed features $\{\hat F_i\}$, we further apply an average-pooling operation to obtain a final feature vector for further processing, which can be summarized as follows:
\begin{equation}
{Y_{fused} = avg(\hat F_1,\hat F_2,...,\hat F_8)}
\end{equation}

Note that we adopt the average-pooling operation as it achieves better performance than the max-pooling operation in our experiment.

\subsection{Embedded into Deep Architectures}
\begin{table*}[htb]
	\caption{Comparisons of the classification accuracy (\%) for rotation robustness on ModelNet40 \cite{wu20153d} classification benchmark using the same test standard as Spherical CNN \cite{esteves2018learning}. \label{tab:m40}
	}
	\centering
	{\small \def\arraystretch{1}\setlength\tabcolsep{10pt}
		\begin{tabular}{ccccccc}
			\hline
			Method & input & input size & z/z & SO3/SO3 & z/SO3  \\
			\hline
			SubVolSup MO \cite{qi2016volumetric} & voxel & $30^3$ & 89.5 & 85.0 & 45.5  \\
			Spherical CNN \cite{esteves2018learning} & projected voxel & $2 \times 64^2$ & 88.9 & 86.9 & 76.7 \\
			\hline
			MVCNN 80x \cite{su2015multi} & views & $80 \times 224^2$ & 90.2 & 86.0 & 81.5  \\
			RotationNet 20x \cite{kanezaki2018rotationnet}& views & $20 \times 224^2$  & \textbf{92.4} & 80.0 & 20.2  \\
			\hline
			PointNet \cite{qi2017pointnet} & xyz & $2048 \times 3$ & 89.2 & 83.6 & 14.7  \\
			PointNet++ \cite{qi2017pointnet++} & xyz & $1024 \times 3$ & 89.3 & 85.0 & 28.6  \\
			SFCNN \cite{rao2019spherical} & xyz & $1024 \times 3$ & 91.4 & \textbf{90.1} & 84.8 \\ 
			ClusterNet \cite{chen2019clusternet} & xyz & $1024 \times 3$ & 87.1 & 87.1 & 87.1 \\
			DGCNN \cite{wang2018dynamic} & xyz & $2048 \times 3$ & 91.9 & 88.3 & 37.8 \\	
			DGCNN (without STN) \cite{wang2018dynamic} & xyz & $2048 \times 3$ & 91.6 & 88.1 & 36.3 \\
			\hline
			Ours (single-frame)  & xyz & $2048 \times 3$ & 89.1 & 89.1 & 89.1 \\ 
			Ours (multi-frame)  & xyz & $2048 \times 3$ & 89.8 & 89.8 & \textbf{89.8} \\ 
			\hline
		\end{tabular}
	}
\end{table*}
\begin{table*}[htb]
	\caption{Comparisons of our approach with state-of-the-art methods on the perturbed dataset of the SHREC’17 \cite{savva2017large} contest. The score, i.e the average of micro and macro mAP is used for ranking and we evaluate our method by using 2048 points as our model's input.}
	\centering
	{\small\def\arraystretch{1}\setlength\tabcolsep{4.5pt}
		
		\begin{tabular}{c|ccccc|ccccc|c}
			\hline
			& \multicolumn{5}{|c|}{micro} & \multicolumn{5}{c|}{macro} &                                 \\
			Method & P@N     & R@N  & F1@N  & mAP & NDCG  & P@N     & R@N  & F1@N  & mAP & NDCG  & score  \\
			\hline
			Furuya \cite{furuya2016deep}   & \textbf{0.814}  & 0.683 & 0.706 & 0.656 & 0.754 & 0.607 & 0.539 & 0.503 & 0.476 & 0.560 & 0.566  \\
			Tatsuma \cite{tatsuma2009multi} & 0.705  & \textbf{0.769} & 0.719 & 0.696 & 0.783  & 0.424 & 0.563 & 0.434 & 0.418 & 0.479  & 0.557 \\
			
			Zhou \cite{bai2016gift} & 0.660  & 0.650 & 0.643  & 0.567 & 0.701 & 0.443 & 0.508 & 0.437  & 0.406 & 0.513    & 0.487  \\
			\hline
			Spherical CNN \cite{esteves2018learning} & 0.717 & 0.737 & - & 0.685 & - & 0.450 & 0.550 & - & 0.444 & - & 0.565 \\
			SFCNN \cite{rao2019spherical} & 0.778 & 0.751 & \textbf{0.752} & 0.705 & 0.813 & 0.656 & 0.539 & 0.536 & 0.483 & 0.580 & 0.594 \\
			DGCNN (without STN) \cite{wang2018dynamic} & 0.768 & 0.717 & 0.719 & 0.672 & 0.782 & 0.640 & 0.527	& 0.515 & 0.449 & 0.564 & 0.561 \\
			DGCNN \cite{wang2018dynamic} & 0.774 & 0.723 & 0.725 & 0.679 & 0.789 & 0.640 & 0.531 & 0.521 & 0.454
			& 0.567 & 0.567 \\
			\hline
			Ours (single-frame) & 0.789 & 0.738 & 0.739 & 0.703 & 0.803 & 0.671 & 0.546	 & 0.539 & 0.479 &	0.585
			& 0.591 \\
			Ours (multi-frame) & 0.801 &	0.747 &  0.749 &	 \textbf{0.714} &  \textbf{0.814} &	\textbf{0.679} &	\textbf{0.563}	& \textbf{0.553} &	\textbf{0.495} &	\textbf{0.601}
			& \textbf{0.605} \\
			\hline
		\end{tabular}
	}
	\label{tab:shrec}
\end{table*}

As we have claimed, our method can be flexibly embedded into the current neural architectures. In this part, we adopt DGCNN \cite{wang2018dynamic} as our basic architecture and further demonstrate how to endow it with rotation-invariance.

The extended architecture, depicted in Figure 4, consists of four modules: PCA-RI representation module, EdgeConv module, self-attention module and classification module. The EdgeConv module contains eight EdgeConv blocks, which share the same weight parameters. Each block consists of five layers with layer output size 64, 64, 64, 128, 1024 respectively. Since our PCA-RI can maintain rotation-invariance, we remove the spatial transform network (STN) of DGCNN \cite{wang2018dynamic} as STN is mainly designed to make model resistant to affine transformation.

For each input point cloud, we first convert it into eight PCA-RI representations and feed them to the eight EdgeConv blocks during the training phase. These blocks further produce eight output features, which will be aggregated by a self-attention module followed by an average pooling layer to  obtain a final feature for the downstream tasks. For simplicity, we denote this model as our multi-frame model.

Another possible architecture is to apply only one EdgeConv block and meanwhile remove the self-attention module and the average-pooling layer with other parts unchanged. We call this single-frame model. During the training phase, the PCA-RI module randomly selects one of the eight representations as input, which can also improve the the rotation robustness of models as can be seen in the following experiments.

Note that our approach doesn't need to apply any rotation augmentation on the training data, which extremely reduces the computational burdens.

\section{Experiments}

In this section, ModelNet40 \cite{wu20153d} is used as the benchmark for 3D classification task. Next, we conduct experiments on ShapeNet Core55 \cite{chang2015shapenet} for the retrieval task. In the end, we provide some ablation analysis of our approach.
\subsection{ModelNet 3D Shape Classification}
We first evaluate the rotation robustness of our proposed method on ModelNet40 \cite{wu20153d} benchmarks for 3D classification task and further compare our method with other state-of-the-art 3D shape classification models. 
\subsubsection{Data}ModelNet40 \cite{wu20153d} is used as the benchmark for 3D classification tasks. ModelNet40 \cite{wu20153d} dataset consists of 12,311 CAD models from 40 manmade object categories. We use the standard split following PointNet \cite{qi2017pointnet} where 9,843 is used for training and 2,468 is used for testing. Since each CAD model in ModelNet40 \cite{wu20153d} is composed of many mesh faces, we sample 2,048 points from them uniformly with respect to face area and then shift and normalize each point cloud into $[-1, 1]^3$ with centroid on the origin. Only the (x, y, z) coordinates of the sampled points are used and the original meshes are discarded.  
\subsubsection{Results} Following Spherical CNN \cite{esteves2018learning}, we evaluate our model using three different settings: 1) training and testing with azimuthal rotations (z/z), 2) training and testing with arbitrary rotations (SO3/SO3), and 3) training with azimuthal rotations while testing with arbitrary rotations (z/SO3).

Table 1 shows the comparisons between our proposed method and the previous methods. All competing methods using azimuthal rotations augmentation suffer a sharp drop on the arbitrary rotation-augmented test set, even for the SO(3) equivariant method Spherical CNN \cite{esteves2018learning} (2 \% and 12.2 \% drop in SO3/SO3 and z/SO3 respectively) and SFCNN \cite{rao2019spherical} (1.3 \% and 6.6 \% drop in SO3/SO3 and z/SO3 respectively) while our approach consistently maintains superior performance across different settings. 
Furthermore, it illustrates that rotation-augmentation can indeed improve the rotation robustness of models but still has a large margin with our proposed method and SFCNN \cite{rao2019spherical} on the SO(3)/SO(3) setting.

Note that SFCNN \cite{rao2019spherical} can  achieve 0.3\% better performance than ours on the SO3/SO3 setting. Nevertheless, SFCNN \cite{rao2019spherical} has to apply a complicated operation to project the point cloud onto a fractal structure, which might  lead to loss of information of the original point cloud. Given the rather simple architecture of our model and the information-lossless input representation we use, we interpret our performance as strong empirical support for the effectiveness of our method.
\subsection{SHREC’17 3D Shape Retrieval}
We also conduct 3D shape retrieval experiments on ShapeNet Core55 \cite{chang2015shapenet} benchmark using  its perturbed dataset, which contains random SO(3) arbitrary rotations. 
\subsubsection{Data} ShapeNet Core55 \cite{chang2015shapenet} benchmark has two evaluation datasets: normal and perturbed. For normal dataset, all model data is consistently aligned while in the perturbed dataset each model data has been randomly rotated by a uniformly sampled rotation. In order to validate the rotation robustness of our approach, we only consider the perturbed dataset which contains a total of 51,190 3D models with 55 categories. 70\% of the dataset is used for training, 10\% for validation, and 20\% for testing.
\subsubsection{Results} Following the experimental settings in Spherical CNN \cite{esteves2018learning}, we train the classification model on the 55 core classes with joint supervision of triplet loss and softmax loss. We use the output of the layer before the score prediction layer as our feature vector and  compute the distance between samples by cosine similarity. 

SHREC'17 \cite{savva2017large} provides several evaluation metrics including Precision, Recall, F1, mAP and normalized discounted cumulative gain (NDCG). 
These metrics are computed under both micro and macro context. 
We evaluate our method and compare it to the prior models using the official metrics. 
In addition, following \cite{savva2017large} we use the average of the micro and  macro mAP as the final score to rank the performance. 

In Table 2, comprehensive comparisons between our approach and various state-of-the-art methods are presented. As we can see, our approach outperforms all other models including the previous state-of-the-art SFCNN \cite{rao2019spherical} under both macro and micro context in terms of most metrics. More importantly, our method is more scalable and flexible without extra complicated operation. 
\subsection{Ablation Analysis}
\subsubsection{Analysis of Architecture} Since our PCA-RI representation can be processed to be compatible with many architectures dealing with point cloud, we further  enhance PointNet \cite{qi2017pointnet++} and PointNet++ \cite{qi2017pointnet++} with our PCA-RI representation. As shown in Table 3, the enhanced DGCNN \cite{wang2018dynamic} (without STN), PointNet \cite{qi2017pointnet} and PointNet++ \cite{qi2017pointnet++} by using the PCA-RI representation outperform the original models  by a large margin  on arbitrary rotation-augmented  dataset for ModelNet40 \cite{wu20153d} classification tasks. 
\begin{table}[htb]
	\centering
	\caption{Analysis of Architecture. These metrics are tested on ModelNet40 \cite{wu20153d} benchmark with 1024 points as input using the SO3/SO3 setting for 3D object classification task. }
	\vspace{6pt}
	\small\def\arraystretch{1}\setlength\tabcolsep{5pt}
	\begin{tabular}[width=\linewidth]{c|c|c}
		\hline
		Method & Model              & Accuracy (\%) \\ \hline
		
		\multirow{3}*{Original Model} & PointNet   & 82.3  \\ 
		~ & PointNet++   & 85.0 \\
		~ & DGCNN (without STN)   & 87.4 \\ \hline
		\multirow{3}*{Ours (single-frame)} & PointNet   &  85.7 \\ 
		~ & PointNet++   & 87.4 \\
		~ & DGCNN (without STN)& 88.2 \\ \hline
		\multirow{3}*{Ours (multi-frame)} & PointNet   &  86.2\\ 
		~ & PointNet++   & 87.9 \\
		~ & DGCNN (without STN)& 88.8 \\ \hline
	\end{tabular}
	
	\label{tab:architecture}
\end{table}
\subsubsection{Analysis of Self-Attention Module}
For our multi-frame approach, how to aggregate the features  for all intrinsic frames is important to extract a discriminative feature for further processing. As shown in Table 4, our experimental results  demonstrate that the performance of directly applying a pooling layer on all the individual features can be improved by adding a self-attention module before the pooling layer. In addition, it illustrates that the average-pooling scheme is a more robust and beneficial to our classification tasks compared with using a max-pooling operation.

\subsubsection{Analysis of Frame Stability}
An important requirement for our approach is that the intrinsic frame should be stable. For an identical object, we hope that the intrinsic frame derived from different sampled point clouds  will maintain consistent. To this end, we make a statistic on the stability regarding the sampling. Our experiment further shows that the average rotation angles between the intrinsic frames derived from two sampled point clouds of the same mesh are $9.52^{\circ}$ and $8.23^{\circ}$ for ModelNet40 \cite{wu20153d} and ShapeNet Core55 \cite{chang2015shapenet} respectively, indicating that the sampled point clouds can have consistent intrinsic frames. 

As aforementioned, another influential factor of our frame stability is the axis significance, which is related to how different the eigenvalues are. We list the distributions of eigenvalue ratio in Figure 5 for ModelNet40 \cite{wu20153d} and ShapeNet Core55 \cite{chang2015shapenet}. Our experimental results demonstrate that more than eighty percent of the point clouds have significant axis order with the ratio of $\frac{\lambda_3}{\lambda_2}, \frac{\lambda_2}{\lambda_1} (\lambda_1>\lambda_2>\lambda_3 )$ being smaller than 0.8. 

\section{Conclusion}
\begin{table}[tb]
	\centering
	\caption{Comparison of different feature fusion methods. These metrics are tested on ModelNet40 \cite{wu20153d} benchmark using DGCNN (without STN) as our backbone network on SO3/SO3 setting for 3D object classification task.}
	\vspace{6pt}
	\def\arraystretch{1}\setlength\tabcolsep{5pt}
	\begin{tabular}[width=\linewidth]{c|c|c}
		\hline
		Input Size & Method              & Accuracy(\%) \\ \hline
		
		\multirow{4}*{$1024 \times 3 $} & Max Pooling   & 87.9 \\ 
		~ & Avg Pooling   & 88.4 \\
		~ & Self Attention + Max pooling   & 88.5 \\
		~ & Self Attention + Avg pooling   & 88.8 \\ \hline	
		
		\multirow{4}*{$2048 \times 3 $} & Max Pooling   & 89.1 \\ 
		~ & Avg Pooling   & 89.5\\
		~ & Self Attention + Max pooling  & 89.5 \\
		~ & Self Attention + Avg pooling  & 89.8 \\ \hline
		
	\end{tabular}
	
	\label{tab:feature fusion}
\end{table} 

\begin{figure}[t!]
	\centering
	\small
	\subfigure[]{\label{fig_potential_field}\includegraphics[width=0.23\textwidth,height = 0.1555\textheight]{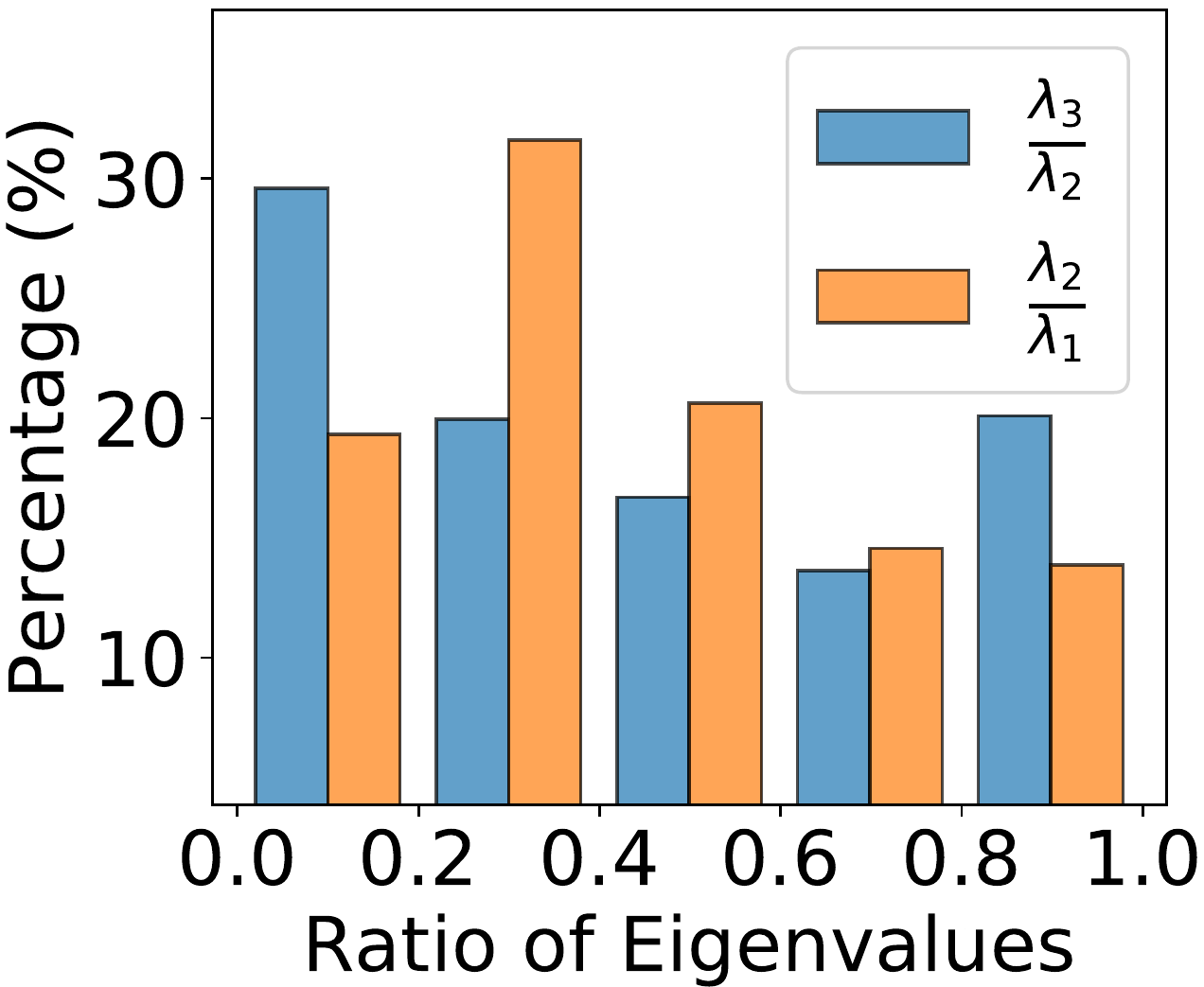}}
	\subfigure[]{\label{fig_som_result}\includegraphics[width=0.23\textwidth,height = 0.1555\textheight]{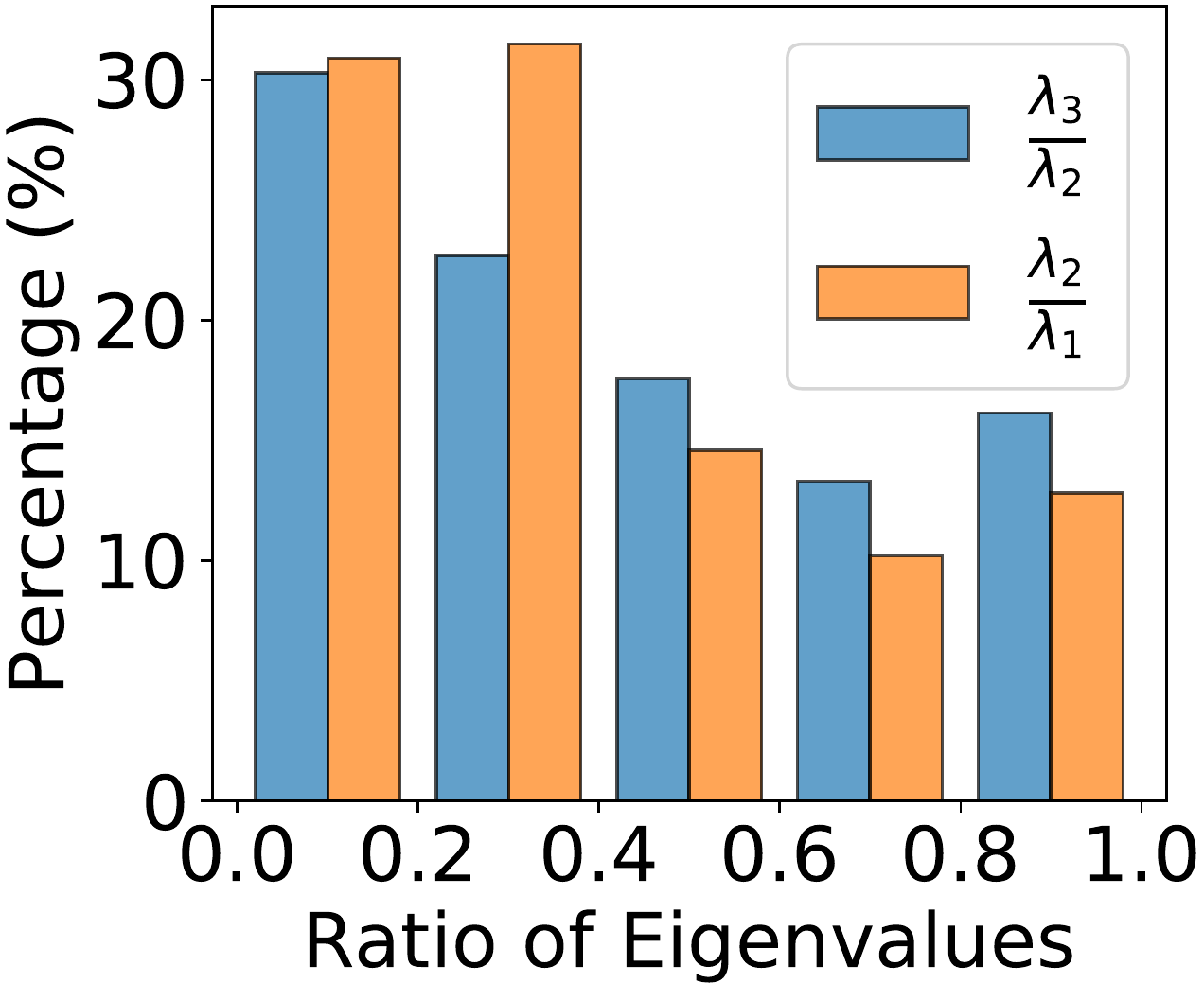}}
	\caption{(a) Distribution of eigenvalue ratios on ModelNet40 \cite{wu20153d}. (b) Distribution of eigenvalue ratios on ShapeNet Core55 \cite{chang2015shapenet}. It illustrates that most of the objects have significant axis order.}
	\label{tab:effects}
	\vspace{-6pt}
\end{figure}
In this paper, we introduce a rotation-invariant representation based on principal component analysis to enhance the rotation robustness for 3D deep models. In order to handle the sign ambiguity of eigenvectors, we adopt a multi-frame  strategy to aggregate all the feature vectors by self-attention mechanisms, which can still preserve the property of rotation invariance theoretically while achieving better performance than directly pooling. Despite its simplicity, our approach is very effective and can be easily embedded to 3D deep models. Extensive experimental results on ModelNet40 and ShapeNet Core55 benchmark demonstrate the superiority of our novel representation.

\bibliographystyle{aaai}
\bibliography{aaai2020}

\end{document}